\pdfoutput=1

\documentclass[11pt]{article}

\usepackage{acl}

\usepackage{times}
\usepackage{latexsym}

\usepackage[T1]{fontenc}

\usepackage[utf8]{inputenc}

\usepackage{microtype}
\usepackage{inconsolata}
\usepackage{makecell}
\usepackage{CJKutf8}
\usepackage{booktabs}	
\usepackage{graphicx}
\usepackage{enumerate}
\usepackage{url}
\usepackage{multirow}
\title{Incomplete Utterance Rewriting as Sequential Greedy Tagging}

\author{Yunshan Chen \\
  SF Technology Co., Ltd. \\
  \texttt{chenyunshan.ai@gmail.com}}

\begin{document}
\maketitle
\begin{abstract}
The task of incomplete utterance rewriting has recently gotten much attention. Previous models struggled to extract information from the dialogue context, as evidenced by the low restoration scores. To address this issue, we propose a novel sequence tagging-based model, which is more adept at extracting information from context. Meanwhile, we introduce speaker-aware embedding to model speaker variation. Experiments on multiple public datasets show that our model achieves optimal results on all nine restoration scores while having other metric scores comparable to previous state-of-the-art models.
Furthermore, benefitting from the model's simplicity, our approach outperforms most previous models on inference speed.
\end{abstract}

\section{Introduction}

Recent years have witnessed increasing attention
in dialogue systems mainly due to its promising
potential for applications like virtual assistants or
customer support systems \cite{hauswald2015sirius,debnathidentifying}. 
However, studies \cite{carbonell1983discourse} show
that users of dialogue systems tend to use incomplete utterances which usually omit (a.k.a. ellipsis) or refer back (a.k.a. co-reference) to the concepts that appeared in previous dialogue contexts. (also known as non-sentential utterances, \cite{fernandez2005using}.
Thus, dialogue systems must understand these incomplete utterances to make appropriate responses.

To tackle the problem, the task of \textbf{I}ncomplete \textbf{U}tterance \textbf{R}ewriting(\textbf{IUR}, also known as context rewriting) \cite{iur-su2019improving,iur-pan2019improving,iur-elgohary2019can}, which aims to rewrite an incomplete utterance into an utterance that is semantically equivalent but self-contained to be understood without context, has recently become an increasing focus of NLP research.
As depicted in Table \ref{tb:iur_example}, the incomplete utterance u3 not only omits the subject
\begin{CJK*}{UTF8}{gbsn} “深圳”\end{CJK*}(Shenzhen), but also refers to the semantics of
\begin{CJK*}{UTF8}{gbsn}
“经常阴天下雨”\end{CJK*}(always raining) via the pronoun \begin{CJK*}{UTF8}{gbsn}
“这样”\end{CJK*}(this).
The downstream dialogue model only needs to take the last utterance by explicitly recovering the dropped information into the latest utterance. Thus, the burden of long-range reasoning 
can be primarily relieved, making the downstream 
dialogue modeling more accurate. 


\begin{table}[]
\centering
\resizebox{\linewidth}{!}{
\begin{tabular}{cc}
\toprule
\textbf{Turn} &  \textbf{Utterance with Translation} \\
\midrule
$u_1$ (A) &  \makecell[c]{\begin{CJK*}{UTF8}{gbsn}
深圳最近天气怎么样？ 
\end{CJK*}\\
(How is the recent weather in Shenzhen?)} \\
\midrule
$u_2$ (B) &  \makecell[c]{\begin{CJK*}{UTF8}{gbsn}
最近经常阴天下雨。
\end{CJK*}
\\ (It is always raining recently.)} \\
\midrule
$u_3$ (A) &  \makecell[c]{\begin{CJK*}{UTF8}{gbsn}
冬天就是这样的。
\end{CJK*}\\ 
(Winter is like this.)}\\
\midrule
$u_{3}^{\prime}$ &  \makecell[c]{
\begin{CJK*}{UTF8}{gbsn}
深圳冬天就是经常阴天下雨。
\end{CJK*}
\\ (It is always raining in winter Shenzhen.) } \\
\bottomrule
\end{tabular}
}
\caption{ An example dialogue between speaker A and B,
including the context utterances $(u_1, u_2)$, the incomplete utterance $(u_3)$ and the rewritten utterance $(u_{3}^{\prime})$.}
\label{tb:iur_example}
\end{table}

The previous top work on building IUR model mainly includes generation-based methods and tagging-based methods. 
Generation-based solution \cite{iur-su2019improving,iur-pan2019improving,iur-elgohary2019can} consider this task as a standard text-generation problem, adopting a sequence-to-sequence model with
a copy mechanism \cite{iur-seq-gulcehre2016pointing,iur-seq-gu2016incorporating,iur-seq-see2017get}.
However, those methods generate the rewritten utterance from scratch, which introduces an over-large search space and neglects the critical trait that the main structure of a rewritten utterance is always the same as the incomplete utterance.

In order to break through those limitations, tagging-based   approach \cite{iur-tag-liu-etal-2020-incomplete,iur-tag-hao-etal-2021-rast,iur-tag-jin2022hierarchical,iur-tag-zhang2022self,iur-tag-wang2022utterance} was proposed.
For specifically, here we consider models like RUN \cite{iur-tag-liu-etal-2020-incomplete} as a tagging-based method. Its semantic segmentation task can be analogous to the sequence tagging task.
The main difference is that the semantic segmentation task is tagging in two-dimensional coordinates, while the sequence annotation task is tagging in one-dimensional coordinates.

The previous top tagging-based approach generally formalizes IUR as learning the edit operation and corresponding location. The tagging-based approach enjoys a smaller search space than the generation-based approach and can better utilize the information that the main structure of a rewritten utterance is always the same as the incomplete utterance.

Despite their success, existing approach that learning edit operation and the corresponding location has difficulty handling situations where multiple inserts correspond to one position. Moreover, models like RUN adopt a heavy model that takes ten convolution layers in addition to the BERT encoder, which will increase its training time and slows down its infer speed.
More critically, although BERT \cite {devlin-etal-2019-bert} has shown to be powerful in extracting information, the generally low restoration scores prove that previous BERT-based models are ineffective in extracting the information needed for IUR from historical utterances.
Finally, the experimental results of SA-BERT \cite{speaker-aware} demonstrate that explicitly modeling speaker changes has a specific enhancement effect on modeling multi-turn dialogue tasks. The previous approach did not model this critical information.

\begin{figure}
    \centering
    \includegraphics[width=0.5\textwidth]{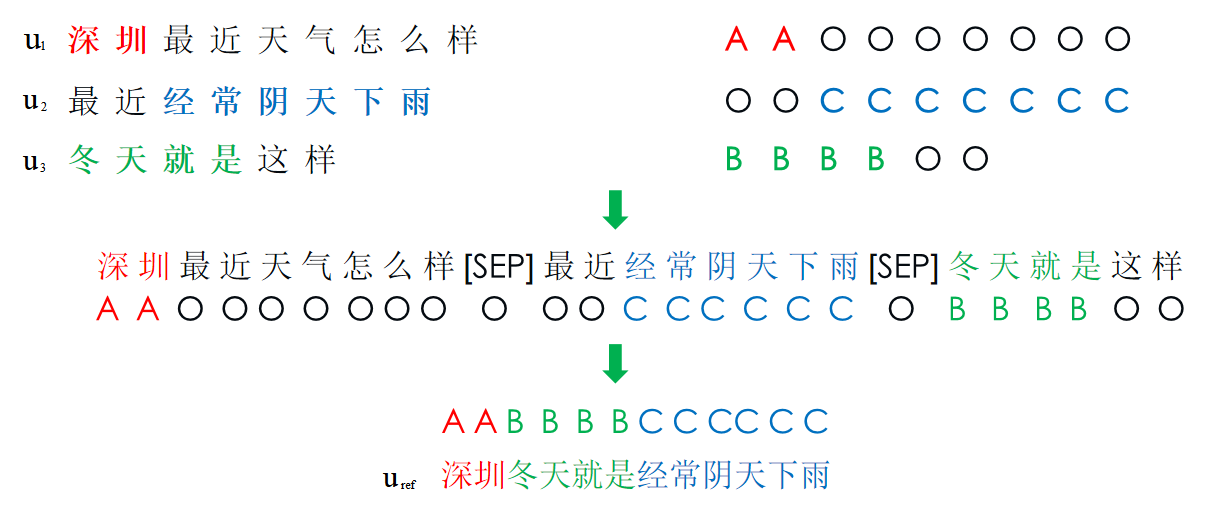}
    \caption{The illustration of the main learning task of the sequence tagging model \textbf{SGT}(\textbf{S}equential \textbf{G}reedy \textbf{T}agging).
    We adopt the same example dialogue from Table \ref{tb:iur_example}.
Considering SGT is position-dependent, and word order between Chinese and English is different, the corresponding English utterance is not provided, which is the same for Figure \ref{fig-sgt-prep-data}.
    }
    \label{fig-sgt-schema}
\end{figure}

To address these issues, we propose a novel sequence tagging model named \textbf{SGT}(\textbf{S}equential \textbf{G}reedy \textbf{T}agging), which is not based on learning editing operations and can significantly improve the restoration score and inference speed.
Our solution was derived from the following thinking: 
First, we consider that in the dialogue process, any complete utterance is composed by only of a few fragments. For example, "I love you" includes three components: subject, verb, and object.
Even if it is expanded with modifications and qualifications, its composition is still minimal.
Based on this insight, we thought it would be possible to build a model to identify the fragments and their order from dialogue history to form a target completed utterance. And then, splice those fragments together in sequence and get the complete utterance.
Meanwhile, in order to keep the number of fragments constituting the target rewritten utterance relatively small, we adopt the greedy tagging strategy. Our model will identify all the fragments and their order required to form a completed utterance; each fragment is the longest fragment found in the given order. We might as well call this fragment \textbf{GLCS} (\textbf{G}reedy \textbf{L}ongest \textbf{C}ommon \textbf{S}ubsequence).
Specifically, we use the tag type to represent the order of GLCS for composing the target rewrite utterance,
For example, the first GLCS that constitutes a rewritten utterance would be tagged as A, and the second is B, the third is C, and so on.
In the above manner, we converted IUR into a simple sequence tagging task, as illustrated in Figure \ref{fig-sgt-schema}.
After the model has identified all GLCSs from the dialogue history through this strategy, the target rewritten utterance can be obtained by splicing each GLCS in alphabetical order according to its tag.


Furthermore, we introduce speaker-aware embedding to model the speaker changes in different rounds. Finally, to better perceive the boundaries of each tagging mention, we add two simple losses in addition to the sequence labeling loss.

In summary, our contributions are as follows:

\begin{enumerate}
    \item We proposed SGT, a novel paradigm to model IUR. Due to the simplicity and effectiveness of modeling, our approach can fully utilize the sequence labeling capabilities of BERT to extract information from historical utterances and thus restore incomplete utterances with more accuracy. Experiments on several datasets show that our method significantly improved the ability to extract mentions from context, which are argued to be harder to copy by \cite{iur-pan2019improving}.
    \item To the best of our knowledge, we are the first to introduce speaker-aware embedding to model IUR.
    \item Finally, benefit from the model's simplicity. Our inference speed is faster than most previous models.
\end{enumerate}


\section{Related Work}
Earlier efforts \cite{iur-su2019improving,iur-elgohary2019can} treated dialogue utterance rewriting 
as a common text generation problem and integrated seq-to-seq models with copy mechanisms to model this task. 
Later work \cite{iur-pan2019improving,zhou-etal-2019-unsupervised,Huang2021SARGAN} explore
task-specific features for additional gains in performance. For example, \cite{iur-pan2019improving} adopt a pipeline-based method. The idea is to detect
keywords first and then append those words to the
context and adopt a pointer generator that takes
the output of the first step to produce the output.
However, this two-step method inevitably accumulates errors.

\begin{figure*}
    \centering
    \includegraphics[width=0.8\textwidth]{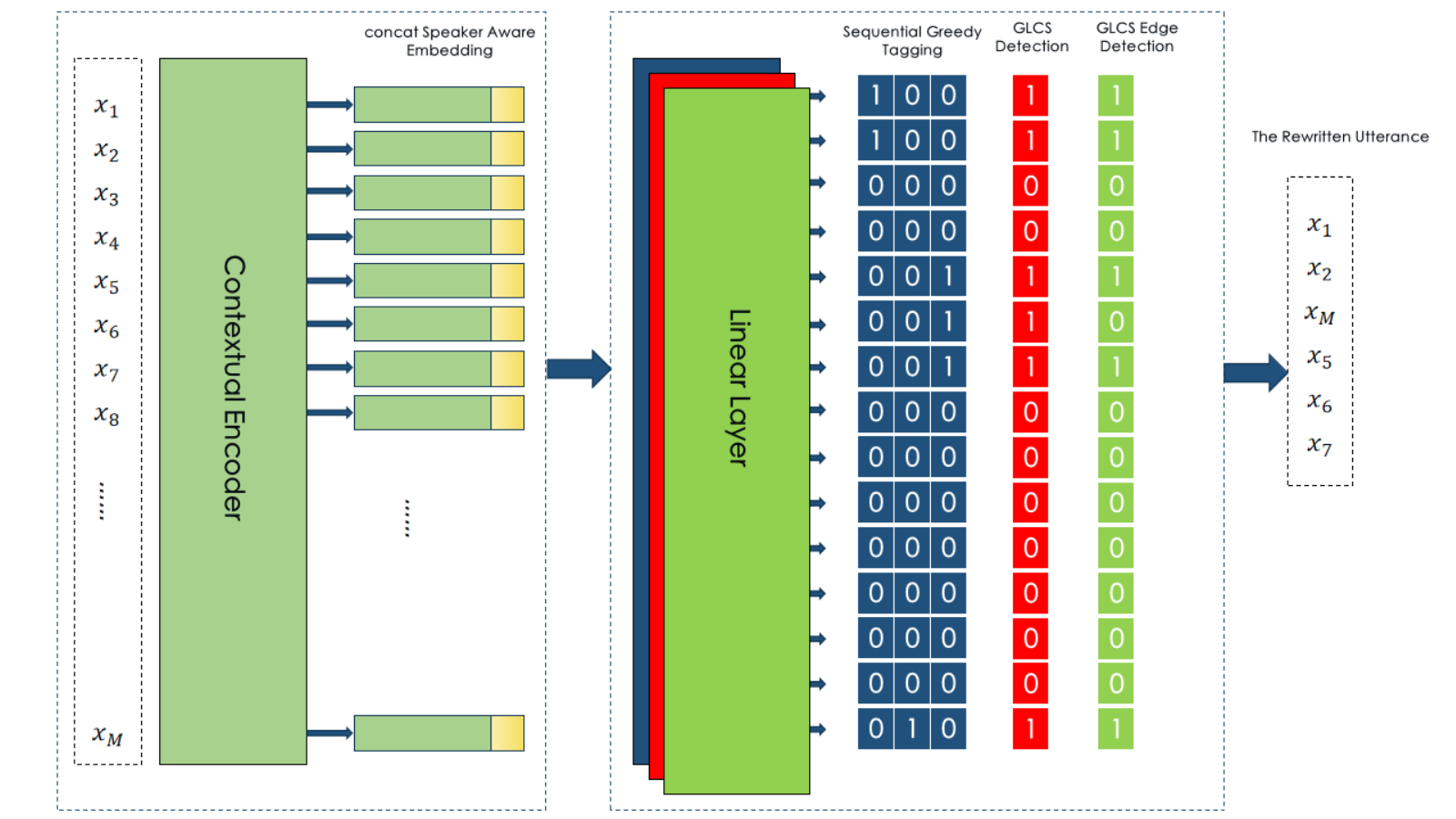}
    \caption{
This figure depicts the SGT model's overall structure, mainly contextual embedding, speaker-aware embedding, 
and three linear layers for the SGT main task and two additional tasks, respectively.
Among the three learning tasks, the dark blue part represents the SGT task, 
the red part represents the \textbf{G}LCS \textbf{D}etection task(GD), 
and the green part represents the \textbf{G}LCS \textbf{E}dge \textbf{D}etection task (GED).
    }
    \label{fig-sgt-arct}
\end{figure*}

SRL \cite{xu-etal-2020-semantic} trains a semantic labeling
model to highlight the central meaning of keywords
in the dialogue as a sort of prior knowledge for the
model. To obtain an accurate SRL model for dialogues, they manually annotate SRL information
for more than 27,000 dialogue turns, which is costly
and time-consuming.

RUN \cite{iur-tag-liu-etal-2020-incomplete} convert this task into a semantic segmentation problem, a significant task in
computer vision. In particular, their model generates a word-level edit matrix, which contains the
operations of insertion and substitution for each
original utterance. Rather than word embeddings, RAU \cite{iur-tag-zhang2022self} 
directly extracts ellipsis and co-reference relationships from Transformer's 
self-attention weighting matrix and edits the original text accordingly to generate complete utterances.  
RUN++ \cite{iur-tag-wang2022utterance} Introduce contrastive learning and keyword detection tasks 
to model the problem jointly. Both RAU and RUN++ make significant improvements
in most metrics on several datasets. Although some
additional effective strategies exist. It is still in the
same paradigm as RUN, learning edit matrix by
cast IUR as a semantic segmentation task.

RAST \cite{iur-tag-hao-etal-2021-rast} is the first work to con-
vert dialogue utterance rewriting into a sequence
tagging task. It takes experimentation to prove that
most models for this task suffer from the robust-
ness issue, i.e., performance drops when testing on a different dataset. 
By contrast, RAST is more robust than the previous works
on cross-domain situations. Moreover, this work 
design additional reinforcement learning task to improve fluency. 
Despite all these efforts, its overall in-domain performance still 
lags behind methods that learn edit operation matric \cite{iur-tag-liu-etal-2020-incomplete}.  

To better enhance pre-trained language models 
for multi-turn response selection in retrieval-based 
chatbots. A model named \textbf{S}peaker-\textbf{A}ware BERT 
(SA-BERT) \cite{speaker-aware} proposed to make 
the model aware of the speaker's changed information, 
which is an essential and intrinsic property of multi-turn dialogues. 

Although RAST has a different learning paradigm from works that learn edit matrix, it still tries to learn the edit operation and corresponding location by sequence tagging. As mentioned before, 
our method is sequence tagging-based but takes an 
entirely new paradigm that would not learn edit 
operations. Besides, inspired by SA-BERT \cite{speaker-aware}, 
we introduce speaker embedding to this task. 
Finally, we introduce two simple sequence labeling tasks to model this problem jointly. 

\section{Methodology}

\subsection{Task Definition}
Here we give the formal definition of how we model the IUR problem with the SGT approach. Taking all history utterances $H = (U_1, U_2, ..., U_n)$  as input, SGT aims to learn a function to rewrite $U_n$ to $R$: $f(H) \rightarrow R$. $R$ is the target rewritten utterance in the infer stage. 
In particular, $U_n$ is the last utterance of all history utterances and the utterance that needs to be rewritten in the IUR task.
$R$ is the reference rewritten utterance $U_{ref}$ in the training phase and the target rewritten utterance in the inference phase.

\subsection{Model Architecture}
Figure \ref{fig-sgt-arct} shows the overall architecture of our model. 

\paragraph{Contextual Encoder}\label{para-encoder}
Since pre-trained language models have been proven to be effective in many NLP tasks, our experiment employs BERT \cite{devlin-etal-2019-bert} to be encoder.
For a fair comparison, we take the same BERT-base encoder as the previous sota work (e.g., RUN, RAU, RUN++) to represent each input. 
Concretely, given input token list $H=\left(x_{1}, x_{2}, \cdots, x_{M}\right)$ which concatenated by all utterances of dialogue history and inserted a special token $[\mathrm{SEP}]$ between each utterance for separate utterances in different turns. The BERT encoder is firstly adopted to represent the input with contextualized embeddings and the calculation of this part is defined as:
\begin{equation}
E = \left(\mathbf{e}_{1}, \cdots, \mathbf{e}_{M}\right)
= BERT\left(H\right)
\label{eq-encoder}
\end{equation}

\paragraph{Speaker Aware Embedding} 
To distinguish utterances between different speakers, 
our approach stitches a one-dimensional one-hot vector at the hidden dimension with the output representation of the BERT encoder. 
This design is based on two considerations. 
On the one hand, most of the dialogue in the dataset is back-and-forth conversations between only two people. 
On the other hand, adding speaker embedding at the input layer and 
performing domain adaptation like SA-BERT will make the encoder different from the BERT-based model, 
which would contradict the fair comparison conditions we assumed earlier in paragraph \ref{para-encoder}. 
The calculation of this part is defined as follows:: 
\begin{equation}
EA=Concat\left(Dropout\left(E\right),SA\right)
\label{eq-sa}
\end{equation}
In the above equation, $E \in R^{M \times 768}$ is the output representation from the contextual encoder. $SA \in R^{M \times S}$ denotes the speaker-aware embedding.
We concatenate $E$ and $SA$ alongside its hidden dimension to get $EA  \in R^{M \times (768+S)}$.

\paragraph{Sequential Greedy Tagging}
Our main task is sequential greedy tagging, this can be generally defined as:  
\begin{equation}P_{sgt}=f(H)\end{equation}
Specifically, ${H}=\left(x_{1}, x_{2}, \cdots, x_{M}\right)$ is the input token list that concatenated by the dialogue's history utterances.
The model learns a mapping function f to predict
from $H$ to the token-level sequence labeling matrix
$P_{sgt} \in R_{MXN}$, where M is the token number of sequence $H$, and N is the number of tag types.
The objective function is defined as:
\begin{equation}
L_{sgt}=\frac{1}{M \times N} \sum_{i=0}^{M \times N} C E\left(P_{sgt}^{i}, Y_{sgt}^{i}\right)
\end{equation}
Where $Y_{sgt}^{i}$ is the target type of the i-th sample at the token level. CE is the notation of cross-entropy loss which is the same for both equations \ref{eq-gd} and \ref{eq-ged}.

\paragraph{GLCS Detection and GLCS Edge Detection}
To better lock in the span of target GLCS needed to make up the rewritten utterance, we introduced multi-task learning. 

Firstly, as depicted by the red components on the right side of Figure \ref{fig-sgt-arct}, the \textbf{G}LCS \textbf{D}etection module (GD) is a binary classification task to distinguish whether a token should belong to a target GLCS. The module GD outputs $P_{gd} \in R_{MX1}$.
LD is essentially a sequence tagging problem, and the loss function of the GLCS detection is as follows:
\begin{equation}
L_{gd}=\frac{1}{M} \sum_{i=0}^{M} C E\left(P_{gd}^{i}, Y_{gd}^{i}\right)
\label{eq-gd}
\end{equation}
$Y_{gd}^{i}$ is the golden mentions label of the i-th sample.
$P_{gd}^{i}$ is the predicted mentions label of the i-th sample.

Secondly, as depicted by the green components on the right side of Figure \ref{fig-sgt-arct}, the \textbf{G}LCS \textbf{E}dge \textbf{D}etection module (GED) is a binary classification task with a structure similar to GD. 
Specifically, a target that consists of a single token or only two tokens will be marked throughout as 1; only tokens at its start position and end position will be
marked as 1 when more than three tokens, left with
the others as 0. 
The loss function of the GED is as follows:
\begin{equation}
L_{ged}=\frac{1}{M} \sum_{i=0}^{M} C E\left(P_{ged}^{i}, Y_{ged}^{i}\right)
\label{eq-ged}
\end{equation}
$Y_{ged}^{i}$ is the golden mentions label of the i-th sample.
$P_{ged}^{i}$ is the predict mentions label of the i-th sample.

\paragraph{Final Learning objectives}
Finally, we combine all tasks and train
them simultaneously by taking the summation of all loss functions, and the final loss function is shown below: 
\begin{equation}
L_{final}=L_{gd}+L_{ged}+L_{sgt}
\label{eq-final-obj}
\end{equation}

\begin{figure}
    \centering
    \includegraphics[width=0.45\textwidth]{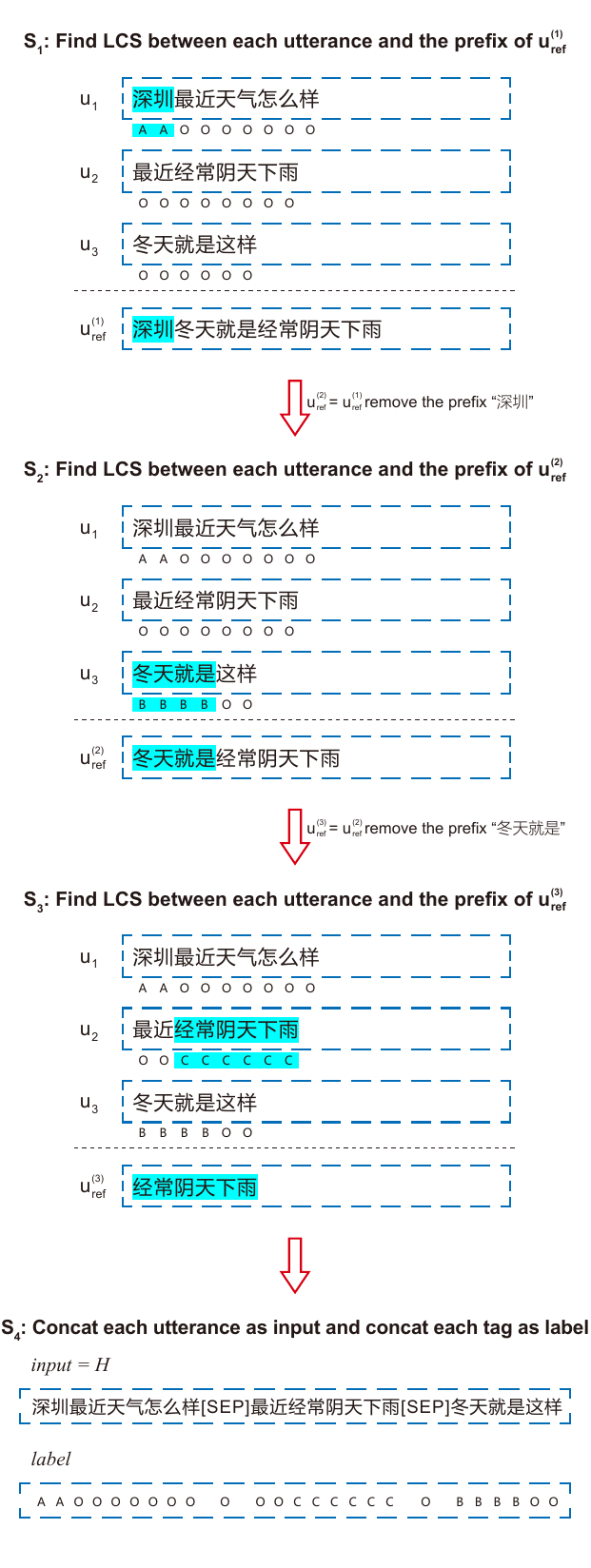}
    \caption{This figure depicts how the training data for the sequence tagging task required by SGT can be generated from the original dataset's history utterances and the reference utterance.}
    \label{fig-sgt-prep-data}
\end{figure}

\subsection{Data Construction}
The construction of the training data for the SGT task is shown in Figure \ref{fig-sgt-prep-data}.
First, in step $S_1$, we make $U_{ref}^{(1)} = U_{ref}$, then find the LCS between each history utterance and $U_{ref}^{1}$ separately. Also, this LCS needs to satisfy being a prefix of $U_{ref}^{(1)}$. After step $S_1$, we can get the first GLCS \begin{CJK*}{UTF8}{gbsn}
“深圳”\end{CJK*}(Shenzhen), and we set the label of its corresponding position to "AA." Then, in step $S_2$, we make $U_{ref}^{(2)}$ = ($U_{ref}^{(2)}$ \textit{remove the prefix} \begin{CJK*}{UTF8}{gbsn}
“深圳”\end{CJK*}(Shenzhen)).
Performing the same GLCS search process, we can obtain the second GLCS \begin{CJK*}{UTF8}{gbsn}
“冬天就是”\end{CJK*}(winter is) and set its label as "BBBB."
Analogously, we can get the third GLCS \begin{CJK*}{UTF8}{gbsn}
“经常阴天下雨”\end{CJK*}(always cloudy and raining) and set its label as "CCCCCC" at Step $S_3$.
Finally, the historical utterances are stitched together as the input of the SGT task. The corresponding labels obtained from steps $S_1$, $S_2$, and $S_3$ are used as the labels of the sequence labeling task.

Points need to be clarified: (i) \textbf{Granularity} The token sequence is char level for Chinese and word level for English and numbers, both in the GLCS matching phase of $S_1$, $S_2$, and $S_3$ and in the training phase of data obtained from $S_4$, which is the same as RUN;
(ii) \textbf{Duplicate Matching} In case of duplicate matches, e.g., if $U_1$ and $U_2$ have the same desired GLCS, the principal is the latter, the better.

\section{Experiments}
In this section, we conduct through experiments to demonstrate the superiority of our approach.

\paragraph{Datasets}
We conduct experiments on three public datasets across different domains: Chinese
datasets in open-domain dialogues: MULTI \cite{iur-pan2019improving} and REWRITE \cite{iur-su2019improving}
, English Task-Oriented Dialogue
TASK \cite{iur-task-quan-etal-2019-gecor}.
 For a fair comparison, We adopt the same data split for these datasets as our baselines. The statistics of these datasets are displayed in Table \ref{tb:data_stat}.
\begin{table}[h]
\centering
\begin{tabular}{c|ccc}
\toprule
 & \textbf{MULTI} & \textbf{REWRITE} & \textbf{TASK} \\
\midrule
Language & Chinese & Chinese & English \\
Train & 194K & 18K & 2.2K \\
Dev & 5K & 2K & 0.5K \\
Test & 5K & NA & NA \\
Avg. C len & 25.5 & 17.7 & 52.6 \\
Avg. Q len & 8.6 & 6.5 & 9.4 \\
Avg. R len & 12.4 & 10.5 & 11.3 \\
\bottomrule
\end{tabular}
\caption{Statistics of different datasets. NA means the
development set is also the test set. “Ques” is short for
questions, “Avg” for short for average, “len” for length, “C”
for context utterance, “Q” for current utterance, and
“R” for rewritten utterance.}
\label{tb:data_stat}
\end{table}

\paragraph{Baselines}
To prove the effectiveness of our approach, we take
the State-of-the-art models as strong baselines including 
SRL \cite{xu-etal-2020-semantic},
SARG \cite{Huang2021SARGAN}, 
PAC \cite{iur-pan2019improving}, 
RAST \cite{iur-tag-hao-etal-2021-rast}, 
T-Ptr-$\lambda$ \cite{iur-su2019improving},
RUN \cite{iur-tag-liu-etal-2020-incomplete} and RUN++ \cite{iur-tag-wang2022utterance}.


\paragraph{Evaluation}
We employ credible automatic metrics to evaluate our approach. As in literature \cite{iur-pan2019improving}, we examine
SGT using the widely used automatic metrics
BLEU, ROUGE, EM and Restoration F-score. (i)
$\textbf{BLEU}_n$ ($\textbf{B}_n$) evaluates how similar the rewritten
utterances are to the golden ones via the cumulative n-gram BLEU score  \cite{eval-Papineni2002BleuAM}.
(ii) $\textbf{ROUGE}_n$ ($\textbf{R}_n$) measures the n-gram overlapping between the rewritten utterances and the
golden ones, while $\textbf{ROUGE}_L$ ($\textbf{R}_L$) measures the
longest matching sequence between them \cite{eval-lin-2004-rouge}. (iii) \textbf{EM} stands for the exact match accuracy, 
which is the strictest evaluation metric.
(iv) $\textbf{Restoration Precision}_n$, $\textbf{Restoration Recall}_n$ and $\textbf{Restoration F-score}_n$
($\mathcal{P}_n$, $\mathcal{R}_n$, $\mathcal{F}_n$) emphasize more on words from dialogue context which are argued to be harder to copy \cite{iur-pan2019improving}. Therefore, they are calculated on the collection of n-grams that contain at least one word
from context utterance. As validated by \citet{iur-pan2019improving}, above
automatic metrics are credible indicators to reflect
the rewrite quality. 

\paragraph{Implementation}
Our implementation was based on PyTorch \cite{impl-pytorch} and fastNLP \cite{impl-fastnlp}. 
In practice, we adopt the exact connection words setting with RUN and append the list of connection words to the head of $H$, as part of it.
Considering that only two speakers are in the datasets related to our experiments, we set the hidden\_size of SA to 1.
For encoding different tagging types, 
We choose \textbf{IO} encoding, the simplest tag encoding schema, which tags each token as either being in (I-X) a particular type of named entity type X or in no entity (O). Since the
distribution of tag types is severely unbalanced
(e.g. (O) accounts for more than 81\% on MULTI), we employed weighted cross-entropy loss and tuned the weight on development sets. We used Adam \cite{impl-adam} to optimize our model and set the
learning rate as 2e-5. We set the dropout rate 
as 0.3 for the dropout operation on the equation \ref{eq-sa}. 
For a fair comparison, the BERT used in our model is BERT-base which is the same as our baselines.

\begin{table*}[h!t]
\centering
\resizebox{\linewidth}{!}{
\begin{tabular}{c|ccccccccccccc}
\toprule
Model & $\mathcal{P}_{1}$ & $\mathcal{R}_{1}$ & $\mathcal{F}_{1}$ & $\mathcal{P}_{2}$ & $\mathcal{R}_{2}$ & $\mathcal{F}_{2}$ & $\mathcal{P}_{3}$ & $\mathcal{R}_{3}$ & $\mathcal{F}_{3}$ & $\textbf{B}_1$ & $\textbf{B}_2$ & $\textbf{R}_1$ & $\textbf{R}_2$ \\
\midrule
SRL & NA & NA & NA & NA & NA & NA & NA & NA & NA & 85.8 & 82.9 & 89.6 & 83.1 \\
T-Ptr-$\lambda$ (n\_beam=5) & NA & NA & 51.0 & NA & NA & 40.4 & NA & NA & 33.3 & 90.3 & 87.7 & 90.1 & 83.0 \\
PAC (n\_beam=5) & 70.5 & 58.1 & 63.7 & 55.4 & 45.1 & 49.7 & 45.2 & 36.6 & 40.4 & 89.9 & 86.3 & 91.6 & 82.8 \\
SARG (n\_beam=5) & NA & NA & 62.3 & NA & NA & 52.5 & NA & NA & 46.4 & 91.4 & 88.9 & 91.9 & \underline{85.7} \\
RAST & NA & NA & NA & NA & NA & NA & NA & NA & NA & 89.7 & 88.9 & 90.9 & 84.0 \\
RUN & 73.2 & 64.6 & 68.8 & 59.5 & 53.0 & 56.0 & 50.7 & 45.1 & 47.7 & \textbf{92.3} & \textbf{89.6} & \underline{92.4} & \underline{85.1} \\
RUN++(PCL) & NA & NA & \underline{71.1} & NA & NA & 59.1 & NA & NA & 51.1 & \underline{92.1} & \underline{89.4} & \underline{92.6} &\textbf{ 86.2} \\
SGT(Ours) & \textbf{75.0} & \textbf{67.5} & \textbf{71.1} &\textbf{ 73.1} & \textbf{65.3} & \textbf{69.0} & \textbf{64.7} & \textbf{57.5} & \textbf{60.9} & \underline{92.1} & \underline{89.0} & \textbf{92.7} & \underline{85.3} \\
\bottomrule
\end{tabular}
}
\caption{Reuslts on MULTI. All models except T-Ptr-$\lambda$ are initalized from pretrained Bert-base-Chinese
model. All results are extracted from the original papers.
The final line is the result of our complete model. A bolded \textbf{number} in a column indicates a sota result against all the other approach, whereas underline \underline{numbers} show comparable performances. Both are same for Table \ref{tb:rewrite_res}\&\ref{tb:task_res}.}.

\label{tb:multi_res}

\end{table*}

\begin{table*}[h!t]
\centering
\begin{tabular}{c|cccccccccc}
\toprule

Model & $\mathcal{F}_{1}$ & $\mathcal{F}_{2}$ & $\mathcal{F}_{3}$ & \textbf{EM} & $\textbf{B}_1$ & $\textbf{B}_2$ & $\textbf{B}_4$ & $\textbf{R}_1$ & $\textbf{R}_2$ & $\textbf{R}_L$ \\
\midrule
SRL & NA & NA & NA & 60.5 & 89.7 & 86.8 & 77.8 & 91.8 & 85.9 & 90.5 \\
RAST & NA & NA & NA & 63.0 & 89.2 & 88.8 & 86.9 & 93.5 & 88.2 & 90.7 \\
RUN & 89.3 & 81.9 & 76.5 & 67.7 & 93.5 & 91.1 & 86.1 & 95.3 & 90.4 & 94.3 \\
RUN++(PCL) & 89.8 & 83.2 & 78.2 & \textbf{69.0} & 93.7 & 91.5 & \textbf{87.0} & 95.6 & \textbf{91.0} & \textbf{94.6} \\
SGT(Ours) & \textbf{91.0} & \textbf{89.8} & \textbf{85.1} & 67.4 & \textbf{94.9} & \textbf{92.2} & \underline{86.8} & \textbf{96.4} & \underline{90.8} & 93.8 \\
\bottomrule
\end{tabular}
\caption{Results on REWRITE. All models are initialized from pretrained Bert-base-Chinese model. All baseline results
are extracted from the RUN++ \cite{iur-tag-wang2022utterance}. The final line is the result of our complete model.}
\label{tb:rewrite_res}

\end{table*}

\begin{table}[h]
\centering
\begin{tabular}{c|ccc}
\toprule
Model & \textbf{EM} & $\textbf{B}_4$ & $\mathcal{F}_{1}$ \\

\midrule
Ellipsis Recovery $^\dagger$ & 50.4 & 74.1 & 44.1 \\
GECOR 1 $^\dagger$ & 68.5 & 83.9 & 66.1 \\
GECOR 2 $^\dagger$ & 66.2 & 83.0 & 66.2 \\
RUN & 70.6 & 86.1 & 68.3 \\
SGT(Ours) & \textbf{71.1} & \textbf{86.7}& \textbf{85.0} \\
\bottomrule
\end{tabular}
\caption{The experimental results on TASK.
$^\dagger$ Results from \citet{iur-task-quan-etal-2019-gecor}.
RUN and SGT are initialized from pretrained Bert-base model, which are same for Table \ref{tb:ab_res}.
}
\label{tb:task_res}
\end{table}
\subsection{Model Comparison}
Table \ref{tb:multi_res}, Table \ref{tb:rewrite_res}, and Table \ref{tb:task_res} show the experimental results of our approach and baselines on MULTI and REWRITE. 
As shown, our approach greatly surpasses all baselines on practically all restoration scores significantly. 
Taking MULTI as an example, our approach exceeds the best baseline RUN++(PCL) on restoration score by a significant margin, reaching a new state-of-the-art performance on almost all restoration metrics. Our approach improves the previous best model by 9.79 points and 9.89 points on restoration $\mathcal{F}_{3}$ and $\mathcal{F}_{2}$, respectively. 
Furthermore, our approach reaches comparable performance on other auto metrics.
As demonstrated by the result of REWRITE, our approach achieves comparable 
performance on the $\textbf{B}_{4}$, $\textbf{R}_{2}$, and $\textbf{R}_{L}$ scores and a new state-of-the-art performance 
on $\textbf{B}_{1}$ and $\textbf{R}_{1}$ scores. Even for the most strict metric EM on REWRITE, 
our approach reached comparable performance with RUN, 
demonstrating the comprehensive ability of our model. 
Besides, our approach achieves better results against 
all baselines on TASK, as depicted in Table \ref{tb:task_res}. 
Specifically, we achieve state-of-the-art performance on the EM score 
and exceed the previous best model by 16.7 points on the restoration $\mathcal{F}_{1}$ score. 
Finally, the combined performance of our model on the three datasets above 
demonstrates that our model can perform well on datasets with varied languages and tasks.

\begin{table*}[h!t]
\centering
\begin{tabular}{l|cccccc}
\toprule
Variant & $\mathcal{P}_{3}$ & $\mathcal{R}_{3}$ & $\mathcal{F}_{3}$ & $\textbf{B}_{1}$ & $\textbf{B}_{2}$ & $\textbf{R}_{L}$ \\

\midrule
SGT & 81.9 & 71.7 & 76.5 & 94.5 & 91.4 & 94.5 \\
SGT w/o (sa) & 80.8 & 71.3 & 75.8 & 94.1 & 91.2 & 94.5 \\
SGT w/o (gd) & 80.9 & 71.6 & 76.0 & 94.1 & 90.9 & 94.4 \\
SGT w/o (ged) & 81.5 & 70.7 & 75.8 & 94.0 & 91.2 & 94.4 \\
SGT w/o (gd+ged) & 82.0 & 68.5 & 74.7 & 93.6 & 90.6 & 94.1 \\
SGT w/o (sa+gd+ged) & 79.4 & 65.5 & 71.8 & 93.6 & 90.3 & 93.8 \\
\midrule
RUN & 70.7 & 45.7 & 55.5 & 91.5 & 89.4 & 93.7 \\
\bottomrule
\end{tabular}
\caption{The ablation results on the development set of TASK.
“SGT” denotes our complete model.
“w/o sa” indicates without the speaker-aware embedding.
“w/o gd” means that remove GLCS detection task from our multi-task learning. 
“w/o ged” means that remove GLCS Edge detection task from our multi-task learning. 
Other remaining variants can be deduced in the same manner.
}
\label{tb:ab_res}

\end{table*}
\subsection{Closer Analysis}
We conduct a series of experiments to analyze our model thoroughly. 
First, we conduct a detailed ablation study to validate the efficacy of the components in our model.
Then, in the same run-time setting, we compare the inference speed of our model to that of representative baselines.
\paragraph{Ablation Study}
By analyzing table \ref{tb:ab_res}, we can find that “w/o sa”,“w/o gd” or “w/o ged” basically hurts the effect of the model, and these can initially corroborate that each of these modules is beneficial to our model.

Meanwhile, we can find that “w/o gd+ged” significantly reduces $\mathcal{R}_{3}$, indicating that these two subtasks are very helpful for discovering the potential target GLCS.
Further, we find that although removing “sa” alone has little effect on the restoration score, comparing the results of removing “gd+ged” and removing “sa+gd+ged” reveals that the fit with the missing speaker-aware information significantly reduces the restoration score. The $\mathcal{F}_{3}$ decreases from 74.7 to 71.8, which indicates that the information of different speakers or rounds is crucial to extract the target GLCS correctly, and combining “sa” embedding and “gd+ged” subtasks can significantly improve the model's ability to obtain the target GLCS fragments from the context.

Finally, we find that even though the absence of the three critical components “sa+gd+gd” leads to an overall decrease in model performance, our model still achieves a better restoration score than the RUN model, which further validates the effectiveness of our sequential greedy tagging learning strategy for modeling and solving UIR problems.

\begin{table}[h] 
\centering
\begin{tabular}{l|ccccc}
\toprule
\textbf{Model} & $\textbf{B}_{4}$ & $\Delta\textbf{B}_{4}$ & \textbf{Latency} & \textbf{Speedup} \\

\midrule
L-Gen & 73.6 & 0.0 & 82 ms & 1.00 × \\
L-Ptr-Gen & 75.4 & +1.8 & 110 ms & 0.75 × \\
T-Gen & 62.5 & -11.1 & 322 ms & 0.25 × \\
T-Ptr-Gen & 77.6 & +4.0 & 415 ms & 0.20 × \\
RUN & 86.2 & +12.6 & 71 ms & 1.15 × \\
\midrule
SGT & 86.8 & +13.2 & 51 ms & 1.60 × \\
\bottomrule
\end{tabular}
\caption{The comparison of inference speeds between SGT and baselines.
We set the beam size parameter to 4 for approaches that need the beam search method, which is not relevant to RUN and SGT.
Meanwhile, we did not do inference performance measurements on RUN++ and other RUN-based models with comparable inference structures, considering they are theoretically nearly identical to RUN.
Latency is calculated as the time it takes to produce a single phrase without data batching, averaged over the REWRITE development set. All models are built in PyTorch and run on a single NVIDIA V100.
}
\label{tb:speed_res}

\end{table}

\paragraph{Inference Speed}
As shown in Table \ref{tb:speed_res}, both SGT and RUN significantly outperform traditional generation algorithms regarding inference speed and $\textbf{B}_{4}$ score. At the same time, the most time-consuming computation of SGT in the inference phase, except for the BERT encoder, is only one layer of a linear transformation, which dramatically saves the inference time compared with RUN, which has  U-net \cite{common-unet} structures after the context encoder. Therefore, we can see that the inference time of SGT is significantly less than that of RUN. The latency of a single rewriting task is reduced by 20ms, while the $\textbf{B}_{4}$ score slightly better.

\section{Conclusion}
In this paper, we convert the IUR problem into a simple sequence tagging task, SGT. The simplicity and effectiveness of the modeling paradigm not only improve the inference speed and allow the pre-trained BERT encoder to fully exploit its widely validated information extraction ability which can significantly improve the restoration score and ensure that other metrics are competitive. We also introduced speaker-aware embedding to explicitly model speaker changes and verified that it has some improvement effect on the IUR task.

In the future, we will explore the following directions:
\begin{enumerate}
\item Adopt the GD task in this paper to extract essential fragments and then pick the best permutation of fragments  with a language model
or using a PAC-like pointer network for fragment integration to get rid of 
the problem of category imbalance is caused by representing the order with tag lists.
\item Combining SGT's efficient fragment extraction paradigm with generation.
\end{enumerate}

\section*{Limitations}
Although our model has made some progress, it still has some limitations. First of all, SGT uses the tag type to represent the connection order of GLCS fragments when forming a complete utterance, and the average statistics on the three datasets we used show that more than 99$\%$ of the complete utterance can be composed with less than three GLCS fragments. That will lead to situations that need to combine multiple GLCSs (e.g., more than 3) to form a complete utterance, which cannot be fully trained or fall into unbalanced tag categories. 
Second, like other tagging-based models, the fragments that make up the complete utterance must exist in history utterances or connection words, which does not work well for situations where it is necessary to combine context information and introduce new words to express their complete utterance.

\section*{Ethics Statement}
We guarantee that the approach in our research is original and that the experimental results of the SGT model are reproducible.  All the experimental data of the baseline model can be reproduced from the relevant open-source code or found in the cited paper.
Finally, The list of authors of the submissions does not include any individuals who did not contribute substantially to work submitted.

\section*{Acknowledgements}
First, we thank all the anonymous reviewers for their valuable comments. Moreover, we are grateful for all the previous work related to exploring the IUR task, which has inspired us a lot.

\bibliography{anthology,custom}
\bibliographystyle{acl_natbib}
\end{document}